\documentclass[letterpaper, 10 pt, conference]{ieeeconf}  

\IEEEoverridecommandlockouts                              

\usepackage{amsmath, amsfonts, amssymb, gensymb}
\usepackage{graphics, graphicx} 
\usepackage{epsfig} 
\usepackage{mathptmx} 

\usepackage{times} 
\usepackage{tabularx}
\usepackage{xcolor}
\usepackage[switch]{lineno}

\newcounter{hypothesis}[section]
\newenvironment{hypothesis}{%
    \refstepcounter{hypothesis}%
    \noindent\textbf{Hypothesis~\thehypothesis.}\ \itshape%
}{\par}

\usepackage{optidef}
\usepackage[english]{babel}

\usepackage{cite}
\makeatletter
\let\NAT@parse\undefined
\makeatother

\usepackage[hyphens]{url}
\usepackage{breakurl}
\usepackage{hyperref}

\title{\LARGE \bf
The Foundational Pose as a Selection Mechanism\\for the Design of Tool-Wielding Multi-Finger Robotic Hands 

}

\author{Sunyu Wang, Jean Oh, and Nancy S. Pollard
\thanks{The authors are with the Robotics Institute at Carnegie Mellon University, Pittsburgh, USA. Corresponding author's contact: sunyuw@andrew.cmu.edu }
}

\begin{document}
\maketitle
\thispagestyle{empty}
\pagestyle{empty}

\begin{abstract}
To wield an object means to hold and move it in a way that exploits its functions. When humans wield tools---such as writing with a pen or cutting with scissors---our hands would reach very specific poses, often drastically different from how we pick up the same objects just to transport them. In this work, we investigate the design of tool-wielding multi-finger robotic hand through a hypothesis: If a hand can kinematically reach a \emph{foundational pose (FP)} with a tool, then it can wield the tool from that FP. We interpret FPs as snapshots that capture the workings of underlying parallel mechanisms formed by the tool and the hand, and one hand can form multiple mechanisms with the same tool. We tested our hypothesis in a hand design experiment, where we developed a sampling-based multi-objective design optimization framework that uses three FPs to computationally generate many different hand designs and evaluate them. The results show that $10,785$ out of the $100,480$ hand designs we sampled reached the FPs; more than $99\%$ of the $10,785$ hands that reached the FPs successfully wielded tools, supporting our hypothesis. Meanwhile, our methods provide insights into the non-convex, multi-objective hand design optimization problem---such as clustering and the Pareto front---that could be hard to unveil with methods that return a single ``optimal" design. Lastly, we demonstrate our methods' real-world feasibility and potential with a hardware prototype equipped with rigid endoskeleton and soft skin. 
\end{abstract}

\section{INTRODUCTION}
Tool-wielding is an essential human skill. It expands our manipulation capability beyond the capability of the biological hand, and is a defining feature of many important jobs centered on physical interaction with the real world \cite{tool_use_survey}. 


Tool-wielding is also different from pick-and-place. Two examples are pens and scissors. When holding a pen to write and holding scissors to cut, our hand would reach very particular poses, which drastically differ from how we grasp the same objects to transport them. Compared with pick-and-place and in-hand re-orientation, tool-wielding with multi-finger robotic hands remains relatively under-explored. 


\begin{figure}[ht]
    \centering
    \includegraphics[width=0.471 \textwidth]{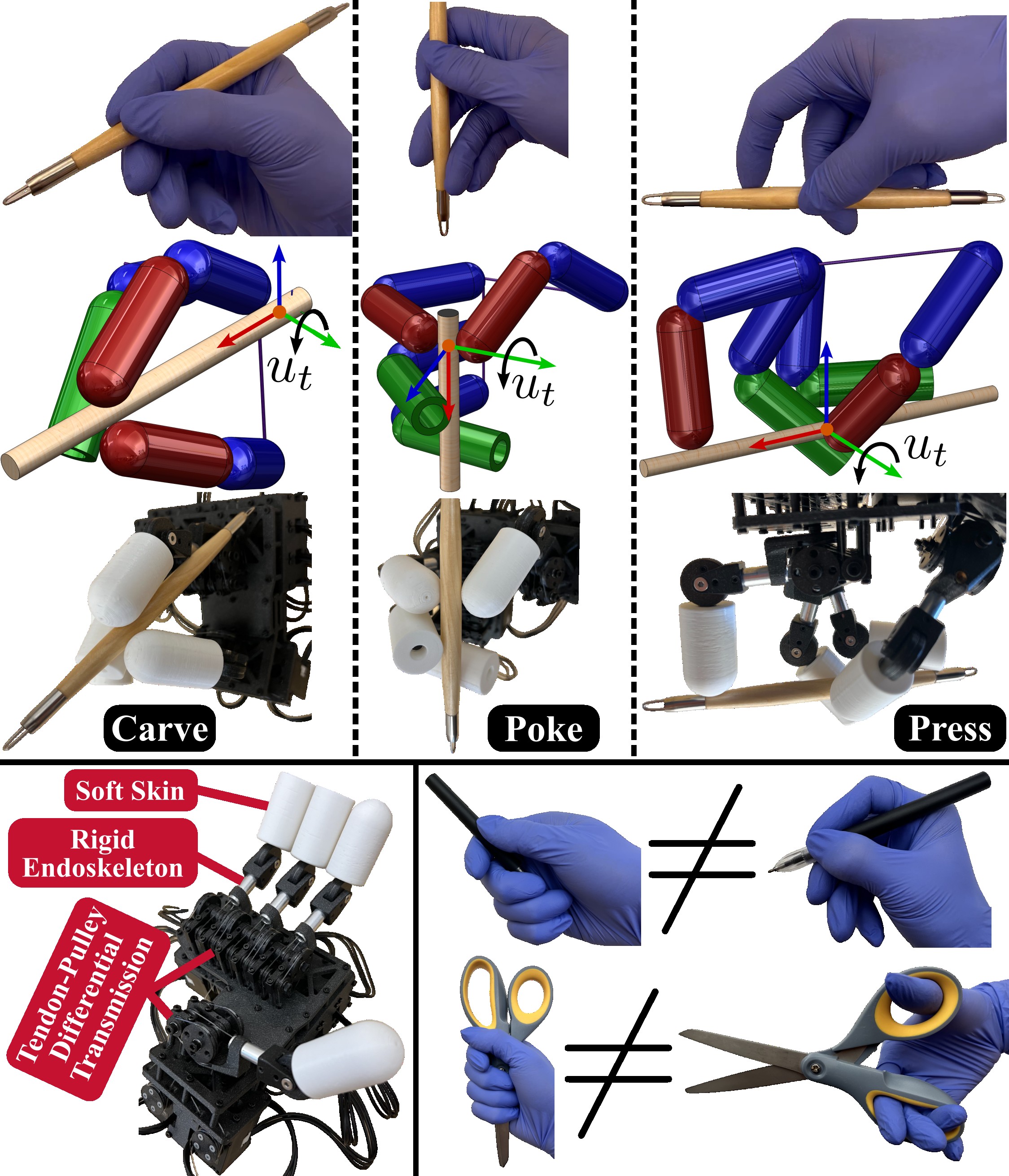}
    \caption{(Top) The three foundational poses identified and implemented in this work. (Bottom left) The hardware prototype. (Bottom right) Examples of grasping an object to transport it vs. wielding a tool. }
    \label{fig:headline}
\end{figure}

In this work, we investigate how to design multi-finger robotic hands that can wield tools. To tackle this problem, we observed how humans wield tools and developed an intuition: Certain poses of the tool-hand system make tool-wielding behaviors easier to analyze. We call these poses \emph{foundational poses (FPs)}. We interpret an FP as a snapshot that captures the working of a parallel mechanism made of the tool and the hand, and one hand can form multiple parallel mechanisms with the same tool. Then, we hypothesize that reaching an FP is a condition for testing if a hand can wield a tool in a specific manner. 

We tested our hypothesis in a hand design experiment, where we developed a sampling-based design optimization framework that uses three FPs to computationally generate and evaluate hand designs. We sampled $100,480$ hands. $10,785$ of them reached all three FPs; more than $99\%$ of the hands that reached all FPs successfully wielded tools, supporting our hypothesis. Furthermore, we built a hardware prototype based on one successful sample and had it perform tool-wielding tasks. Even with open-loop control and under unmodeled conditions, the prototype achieved the tasks. 

This work's primary contribution is the formulation of FPs as a selection mechanism for the design of tool-wielding multi-finger robotic hands. A second contribution is the design framework developed, which generates many designs that permeate the design space, rather than one ``optimal" design. Together, the two contributions demonstrate the potential of the concept of foundational poses for tackling dexterous manipulation problems, and provide insights into the non-convex, multi-objective hand design optimization problem that could be hard to obtain otherwise.

\section{Related Works}
\subsection{Robotic Tool-Wielding}
Existing works on robotic tool-wielding generally come from hand design, robot control, and robot learning communities. Among related works in hand design, industrial automation solutions account for a significant portion \cite{ati_tool_changer, schunk_tool_changer, zimmer_tool_changer, okuma_cnc, hass_cnc, triple_a_robotics, berkshire_grey, da_vinci_surgical_instruments, gitai}. The dominant design philosophy is to make the hand the tool, and to switch between tools via dedicated adaptors and tool changers. These solutions tend to enjoy high practicality and performance, but suffer from bulkiness and poor generalizability due to the specialized and complex hardware. 

For related hand design works outside industrial automation, the dominant design philosophy is anthropomorphism \cite{nature_hand, dlr_david, shadow_hand, ROB_hand_3, soft_hand, hand_using_chopsticks, roboray}, though non-anthropomorphic designs that succeeded in tool-wielding also exist \cite{jam_hand, vacuum_jamming, lip_jamming, high_force_winding}. These works have demonstrated qualitatively impressive and even human-like tool-wielding performances. However, they often regard tool-wielding as a consequent success of the design and a task for showcasing the hardware's capability. Specifically how these hands wield tools and how their design influences their tool-wielding performances are typically not a focus. 

Related works in robot control and robot learning do address how to wield tools in detail. Popular approaches include dynamics-inspired hybrid position and force control \cite{dlr_david_control, screwdriver_2023, screwdriver_2024}, affordance learning \cite{affordance, affordance_icub}, deep reinforcement learning \cite{learning_for_manipulation_survey, robotool, learn_to_design_and_use_tool, hockey, pasta, bear_tool_usage, tool_use_self_supervision}, and learning from demonstration \cite{tool_use_lfd, astribot, tri_diffusion_policy, 3d_diffusion_policy, ldf_affordance}. However, many of these works use parallel jaw grippers, and predominantly reason about the interaction between the tool and the manipulated object. Only a minority of these works explicitly addressed the tool-hand interaction, which may be more critical for tool-wielding multi-finger hands than for parallel jaw grippers. 

\subsection{Computational Design of Robotic Hands}
Computational hand design works formulate hand design as a numerical optimization problem, and leverage computation to explore a large number of design possibilities that are hard for human designers to manually ideate. However, few such works have touched upon tool-wielding. Most of them concern easier-to-formulate tasks, such as pick-and-place and in-hand re-orientation \cite{end_to_end_hand_design_2021, rl_hand_design_2021, hazard_2019, meixner_2019, soft_hand_design_2017}. 

Moreover, hand design optimization problems are often non-convex. When more than one task or evaluation metric is considered, they will become multi-objective. Such problems rarely admit a single design that simultaneously optimizes all objectives. Rather, the set of non-inferior designs---the Pareto front---is a more appropriate definition of optimum. Yet, many existing computational hand design works focus on frameworks that return one ``optimal" design. In essence, this ``optimal" design is a fine-tuned local optimum based on human heuristics, such as initialization, objective function shaping, and regularization. To some extent, these frameworks trade human designers' mechanical design heuristics for their algorithm tuning heuristics. Neither alone is likely to thoroughly uncover the richness of a non-convex, multi-objective hand design optimization problem. 

\subsection{Most Relevant Works}
Our work tackles multi-finger hand design specifically for tool-wielding based on a tool-hand interaction model---the foundational pose---and a sampling-based optimization strategy. For the tool-hand interaction model, our work is similar in spirit to \cite{parallel_hand, stewart_hand, roller_hand, leg_parallel_mechanism}, which treat the robot and the manipulated object or the environment as one parallel mechanism. For the sampling-based optimization strategy, our work is similar in spirit to \cite{soft_hand_design_2017, sampling_design_snake, sampling_design_arm}, which sample many designs and illuminate the performance landscapes with evaluation results of the samples. 

\section{Methods}
Designing tool-wielding multi-finger hands is challenging in general. To scope the problem, we make the following assumptions: 1) The hand will wield zero-degree-of-freedom (0-DoF), rigid clay sculpting tools with known cylindrical geometries. 2) The tools experience inertial forces significantly smaller than the tool-hand and the tool-workpiece interaction forces; or equivalently, the tools have negligible masses. 

\subsection{Motivation and Definition of Foundational Poses}
We observed that human sculptors tend to first hold a tool in a particular pose, and then put the tool in contact with the workpiece. When the tool and the workpiece are in contact, the fingers move in a synchronized and repeatable manner. We interpret these behaviors as follows: After picking up the tool, the human configures the tool-hand system into a parallel mechanism, and moves the fingers in ways that maintain the mechanism's topology \cite{tool-use_2016}. When the parallel mechanism moves, its closed kinematic chains constrain the tool's and the fingers' DoFs, making the fingers' movements predictable once the tool's movement is defined. 


We identified three tool-hand parallel mechanisms used by human sculptors, shown in Fig. \ref{fig:headline}: 1) Carve, a 3-finger mechanism with three contacts arranged in a triangle on the tool's cross section. 2) Poke, a 4-finger mechanism with two pinches arranged perpendicularly along the tool's two non-longitudinal axes. 3) Press, a 4-finger mechanism with a pinch in the tool's middle and two contacts at the two ends. 


The tool exhibits planar rotational movements in all three tool-hand parallel mechanisms. Hence, we define each mechanism's \emph{foundational pose (FP)} as a pose where the tool can rotate in both directions, while satisfying contact constraints with the hand and static equilibrium with an external force experienced at the tool's tip. In other words, an FP is a statically feasible pose that must be passed through if a tool-hand parallel mechanism moves in its full range of motion. This suggests the following 

\vspace{3pt}
\begin{hypothesis} 
\label{hpth:tool_hold}
If a tool and a multi-finger hand can reach a foundational pose, then they can act as a corresponding tool-hand parallel mechanism for the hand to wield the tool and achieve a non-zero tool motion range. 
\end{hypothesis}

This hypothesis is central to this work, as it distills a complex tool-wielding behavior into a single pose. Next, we will test this hypothesis in a hand design experiment. 

\begin{figure}[h]
    \centering
    \includegraphics[width=0.5 \textwidth]{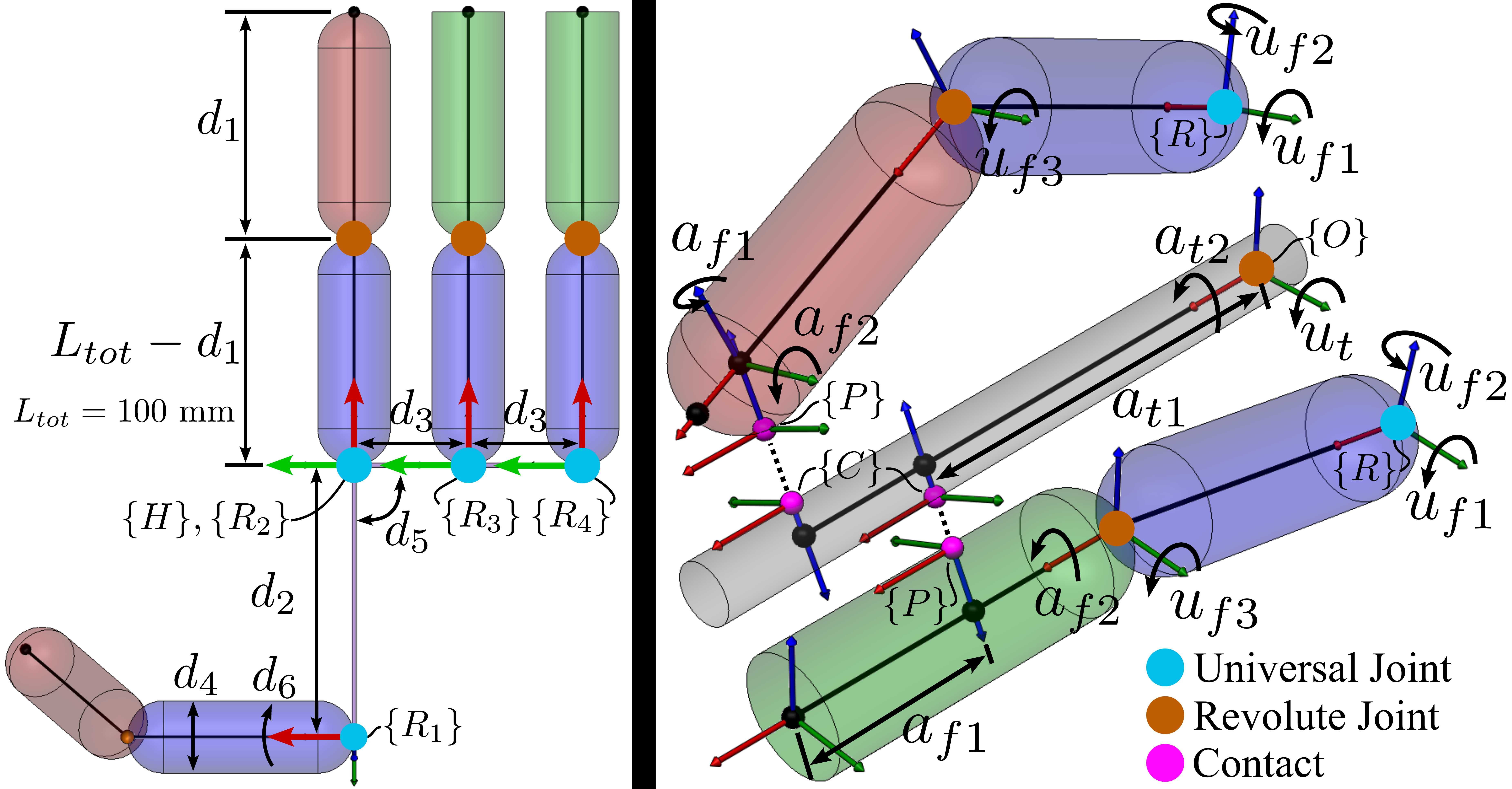}
    \caption{(Left) The design template. (Right) The tool-finger parametrization. The frame axes' colors convention is red-green-blue $\rightarrow xyz$.}
    \label{fig:design_template_and_parametrization}
\end{figure}





\begin{figure*}[t]
    \centering
    \includegraphics[width=0.96 \textwidth]{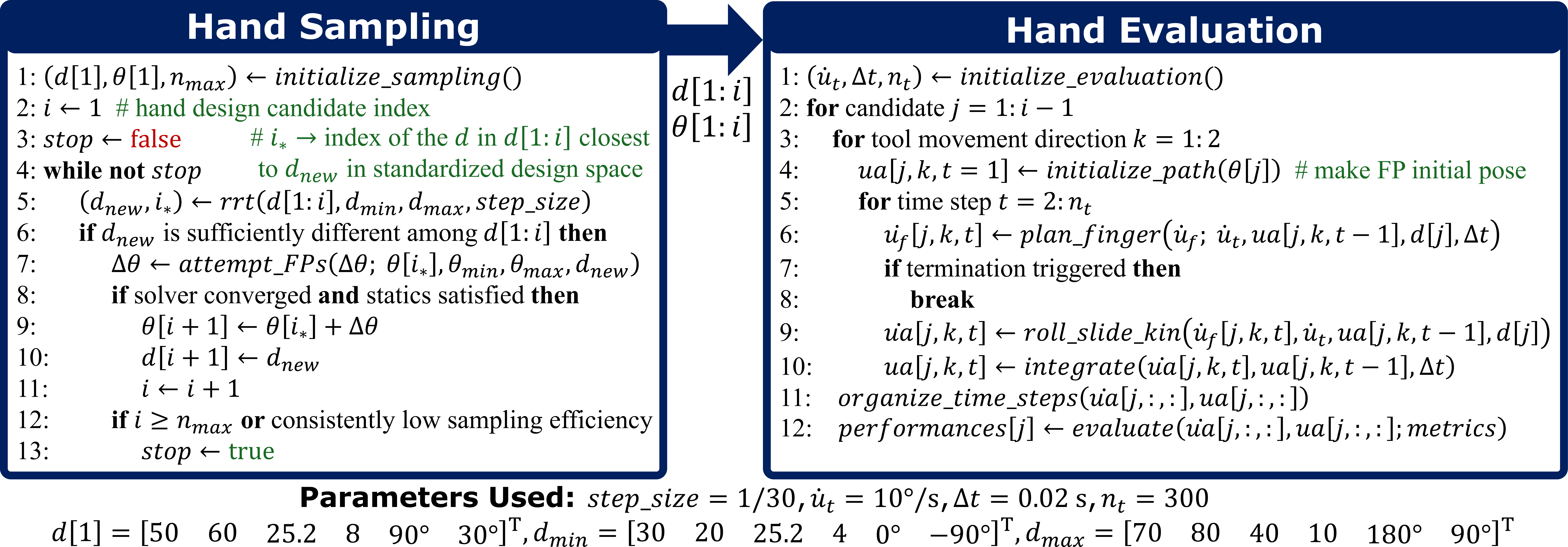}
    \caption{Pseudo-code of the sampling-based hand design optimization framework. Length values are in millimeters.}
    \label{fig:algorithm}
\end{figure*}

\subsection{Design Template and Hardware Realization}
We adopted an anthropomorphic design template to specify the hand's design space. The design template consists of four identical 3-DoF fingers, each with a universal metacarpophalangeal (MCP) joint and a revolute interphalangeal (IP) joint. The MCP joint is actuated via a tendon-pulley differential transmission \cite{barret_wam, romela_bruce}: The finger flexes/extends when the two pulleys rotate in the same direction at equal speeds, and adducts/abducts when the two pulleys rotate in the opposite direction at equal speeds. Each finger also has cylinder- and capsule-shape links, which are realized on hardware with an aluminium rigid endoskeleton and a 3D printed thermoplastic elastomer (TPE) soft skin.  Fig. \ref{fig:design_template_and_parametrization} and Fig. \ref{fig:headline} show the design template and a hardware prototype built based on the template, respectively. The design parameters $d = \begin{bmatrix} d_1 & ... & d_6 \end{bmatrix}^{\intercal}$ define a six-dimensional (6D) design space. Each parameter means the following: 
\begin{itemize}
    \item $d_1 \rightarrow$ the distal link's length. The entire finger's length is $L_{tot} = 100$ mm, so $d_1$ also determines the proximal link's length. 
    \item $d_2 \rightarrow $ the palm's length. 
    \item $d_3 \rightarrow$ the palm's half width. 
    \item $d_4 \rightarrow$ the finger's diameter. 
    \item $d_5 \rightarrow$ the palm's angle. 
    \item $d_6 \rightarrow$ the thumb's axial rotational angle relative to the palm, i.e., rotation about frame $\{ R_{1} \}$'s $x$-axis in Fig. \ref{fig:design_template_and_parametrization}. 
\end{itemize}

We chose this design template and hardware realization because: 1) Four 3-DoF fingers are the fewest fingers and DoFs required for one hand to achieve all three identified FPs. 2) Cylinder and capsule are some of the simplest geometries for reasoning about rolling and sliding contact and collision. 3) The design parameters represent feasible hardware changes in the hardware prototype's mechanical design. 4) The rigid endoskeleton ensures reasonable strength and accuracy of a rigid body kinematics model, while the soft skin provides local compliance at the contact to compensate for modeling errors and uncertainties. 5) The differential transmission amplifies strength along the finger's MCP flexion/extension axis, which is a dominant movement axis in all three identified tool-hand parallel mechanisms. 

\subsection{Tool-Hand System Parametrization}
With the design template, we parametrized the tool-hand system with rigid body kinematics and contact constraints. Specifically, we derived the following forward kinematics functions in the form of homogeneous transformation, as illustrated in Fig \ref{fig:design_template_and_parametrization}: 1) $T_{OC} \left( ua_{t} \right)$, the pose of a moving frame $\{C\}$ on the tool's surface relative to the tool's origin frame $\{O\}$. 2) $T_{RP} \left( ua_{f} \right)$, the pose of a moving frame $\{P\}$ on the surface of a finger's distal link relative to the finger's root frame $\{R\}$. 3) $T_{OH} \left( \theta_{h} \right)$, the pose of the hand frame $\{H\}$ relative to $\{O\}$. The rotation matrices in $T_{OC}$ and $T_{RP}$ are defined as the normalized Gauss frame, so their $z$-axes are inward-pointing surface normals \cite{montana_contact_kinematics, math_intro_to_mani}. $ua_{t} = \begin{bmatrix} u_{t} & a_{t1} & a_{t2} \end{bmatrix}^{\intercal}, ua_{f} = \begin{bmatrix} u_{f1} & u_{f2} & u_{f3} & a_{f1} & a_{f2} \end{bmatrix}^{\intercal}$, and $\theta_{h} \in \mathbb{R}^6$ are the tool's, the finger's, and the hand's states, respectively. $\theta_{h}$ corresponds to a 6-DoF joint. 

The contact constraint for one tool-finger contact is then
\begin{align} \label{eq:contact_constraint}
    p_{OC} = p_{OP}, \quad z_{OC} + z_{OP} = 0,
\end{align}
where $p \in \mathbb{R}^3$ and $z \in \mathbb{R}^3$ are the origin's Cartesian position and the $z$-axis unit vector of the corresponding frame, respectively. Physically, (\ref{eq:contact_constraint}) means that a point on the tool and a point on the finger coincide and their surface normals are anti-aligned. This ensures a point contact without separation or interpenetration between the two contacting bodies. 

(\ref{eq:contact_constraint}) yields six scalar nonlinear equations. Applying $(\ref{eq:contact_constraint})$ to a tool-hand system with $n$ tool-finger contacts yields an under-determined system of $6n$ equations and $\left( 8n + 6 \right)$ variables
\begin{align} \label{eq:maneuver_equation}
    reach\_FP \left( \theta, d \right) = 0, 
\end{align}
where $\theta = \begin{bmatrix} \theta^{\intercal}_{h} & ua^{\intercal}_{t1} & ... & ua^{\intercal}_{tn} & ua^{\intercal}_{f1} & ... & ua^{\intercal}_{fn} \end{bmatrix}^{\intercal}$ is the collection of all states, and $ua_{ti}, ua_{fi}, i \in \left[ 1, \cdots, n \right]$, are the $ua_{t}$ and $ua_{f}$ for tool-finger contact $i$, respectively. We repeat (\ref{eq:maneuver_equation}) with specific bounds of $\theta$ for each of the three FPs. And for carve, $n = 3$; for poke and press, $n = 4$. 


\subsection{Foundational Pose-Constrained Hand Sampling}
The set of solutions to (\ref{eq:maneuver_equation})'s for all FPs defines a manifold in the design space where all hand designs can reach all FPs. We call it the ``feasible design space". To test Hypothesis \ref{hpth:tool_hold}, we aim to sample different hands in this space, and check how many of them can wield the tools. We address this problem with an analogy: The feasible design space's boundary is analogous to a non-penetrable chamber whose shape is unknown. The sampling problem is analogous to estimating the chamber's shape. To do this, we pump gas from one point inside the chamber. By diffusion, the gas molecules gradually permeate the chamber as more gas is pumped, but cannot penetrate the chamber's walls, creating an increasingly better approximation of the chamber's interior shape. 

Each ``gas molecule" is a hand design sample. The ``point inside the chamber" is a manually crafted initial hand design at its FPs. And we use a rapidly exploring random tree (RRT) to sample hand designs from the initial design. In other words, we use the FPs' complex kinematic and contact constraints to restrict the feasible design space. Then, we initialize with one hand known to be inside this feasible design space, and sample more hands from it using RRT. This increases the chance of finding more hands that can reach the FPs compared with, for example, random sampling. 

Fig. \ref{fig:algorithm} left depicts the sampling process. After initialization, the RRT generates a sample based on a uniform distribution and Euclidean distance in the standardized design space, i.e., a design space where elements of $d$ are standardized by the minimum and maximum design parameters, $d_{min}$ and $d_{max}$, to dimensionless values between $-0.5$ and $0.5$. Then, the sample is accepted and becomes a ``candidate" if it 1) sufficiently differs from all other candidates in design parameters, and 2) satisfies (\ref{eq:maneuver_equation}) for all FPs, and 3) satisfies static equilibrium under an external force applied at the tool's tip, which represents the cutting force. This process repeats until either 1) a prescribed number of candidates is reached, or 2) the sampling efficiency, defined as the number of candidates over the number of RRT calls, is consistently too low in a moving average.  


Inside the $attempt\_FPs \left( \cdots \right)$ function on Fig. \ref{fig:algorithm} left line 7 is a nonlinear optimization step, which attempts to move the newly sampled hand to its FPs
\begin{mini*}|s|
{\Delta \theta}{ sampling\_cost\left( \Delta \theta; \theta \left[ i_{*}  \right], d_{new} \right) }
{}{}
\addConstraint{ reach\_FPs \left( \Delta \theta; \theta \left[ i_{*}  \right], d_{new} \right) = 0 }
\addConstraint{ collision \left( \Delta \theta; \theta \left[ i_{*}  \right], d_{new} \right) \geq 0 }{}
\addConstraint{ \theta_{min} - \theta \left[ i_{*}  \right] \leq \Delta \theta \leq \theta_{max} - \theta \left[ i_{*}  \right] },
\end{mini*}
where $d_{new}$ is the newly sampled design parameters, $i_{*}$ is the index of the $d$ in existing candidates that is closest to $d_{new}$ in the standardized design space, and $\theta_{min}$ and $\theta_{max}$ are $\theta$'s bounds specific to each FP. $reach\_FPs\left( \cdots \right) = 0$ is the collection of (\ref{eq:maneuver_equation})'s for all FPs. $collision \left( \cdots \right) \geq 0$ prevents collisions among fingers and between the tool and non-contacting finger links, based on the shortest distance between two line segments \cite{collision_detector}. The arguments after a semicolon are treated as constants in that function. The optimization is performed with the active-set method in MATLAB's fmincon() function \cite{matlab_fmincon}, initialized with zero. 

$sampling\_cost \left( \cdot \right)$ is FP-specific, and is heuristically designed to improve each FP. For carve, the cost is to maximize the fraction of the finger contact force that contributes to the tool's cutting moment. For poke and press, the cost is to maximize the anti-alignment of the two pinching fingers' contact normals. The cost also contains a regularizer that discourages the usage of MCP adduction/abduction, $u_{f2}$, due to its limited motion range. 






\begin{figure}[t]
    \centering
    \includegraphics[width=0.49 \textwidth]{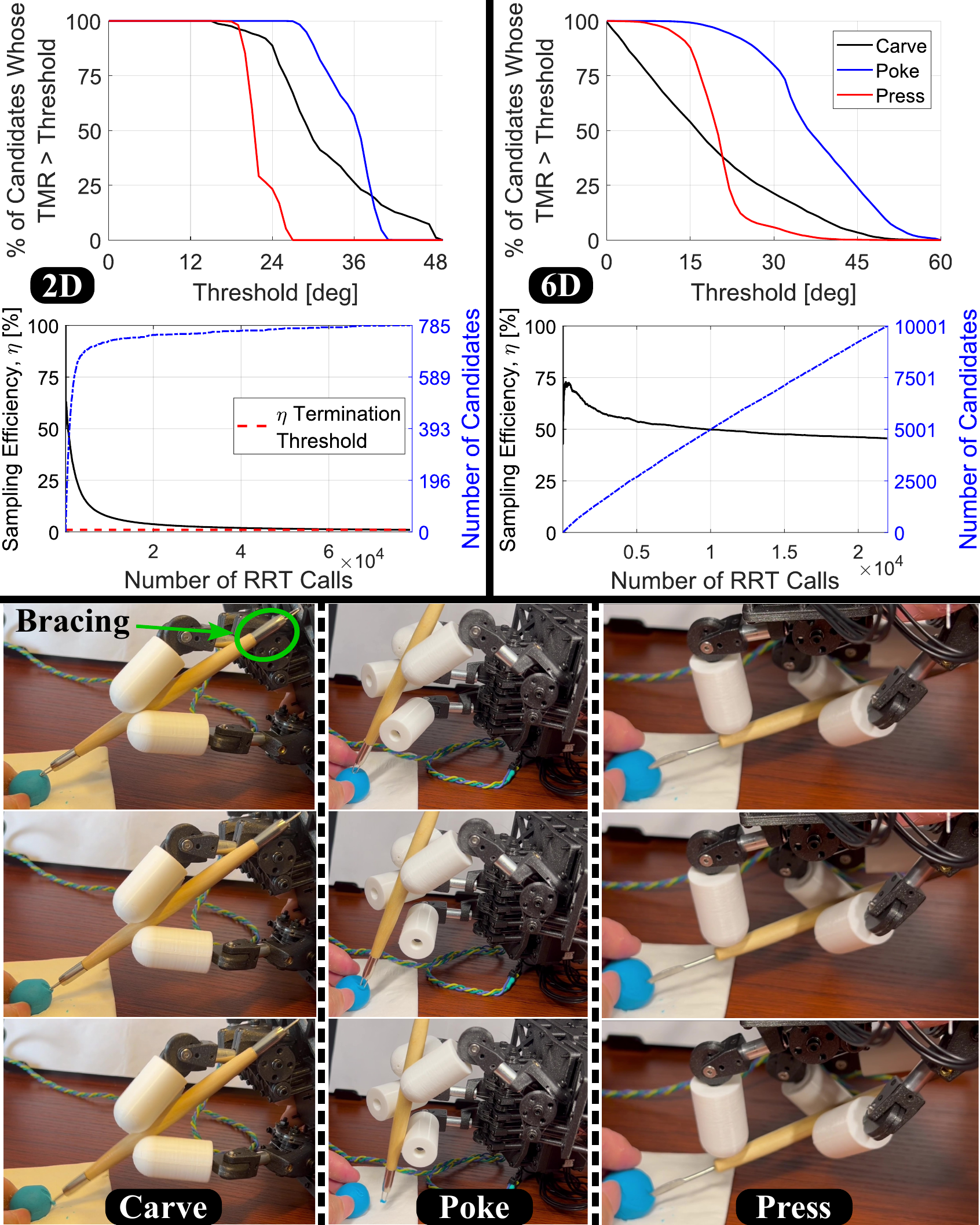}
    \caption{(Top) Sampling efficiency and percentage of candidates whose tool motion ranges are greater than a variable threshold for the 2D (top left) and the 6D (top right) sampling cases. (Bottom) Snapshots of the hardware experiment as the tool contacted the clay. }
    \label{fig:results_eta}
\end{figure}

\begin{figure}[!t]
    \centering
    \includegraphics[width=0.49 \textwidth]{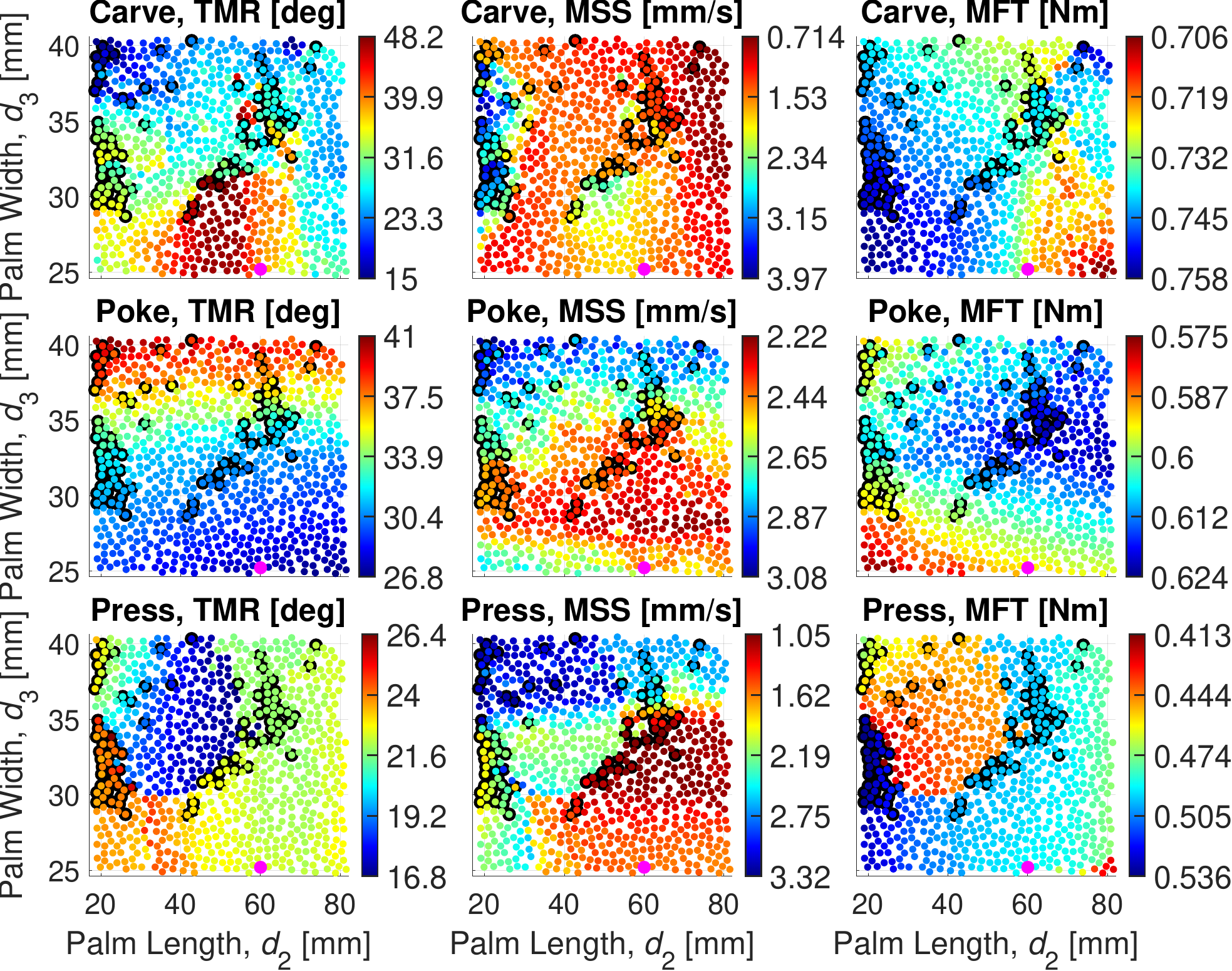}
    \caption{The 2D sampling case's performance landscapes. Each point is a candidate. Redder means better performance. The black-outlined points are non-Pareto-optimal candidates. The magenta point is the candidate based on which the hardware prototype was built. }
    \label{fig:results_2d}
\end{figure}

\subsection{Tool-Wielding Path Planning and Hand Evaluation}
After obtaining the candidates, we seek to evaluate how well they wield tools. To achieve this, we planned each candidate's tool-wielding paths from each FP with the algorithm shown in Fig. \ref{fig:algorithm} right. The $plan\_finger \left( \cdots \right)$ function on line 6 is a quadratic program (QP) that minimizes contact sliding 
\begin{mini*}|s|
{\dot{u}_f}{ v^{\intercal}_{CP} E_{xy}^{\intercal} E_{xy} v_{CP} }
{}{}
\addConstraint{ E_{z} v_{CP} = 0 }
\addConstraint{ motion\_range \left( \dot{u}_f; \dot{u}_t, ua \left[ j, k, t - 1 \right], d \left[ j \right], \Delta t \right) \geq 0 }{}
\addConstraint{ collision\_linearized \left( \dot{u}_f; \dot{u}_t, ua \left[ j, k, t - 1 \right], d \left[ j \right], \Delta t \right) \geq 0 }.
\end{mini*}
As shown in Fig. \ref{fig:design_template_and_parametrization}, $\dot{u}_t \in \mathbb{R}$, $\dot{u}_f = \begin{bmatrix} \dot{u}_{f1} & \dot{u}_{f2} & \dot{u}_{f3} \end{bmatrix}^{\intercal}$. $a_{t} = \begin{bmatrix} a_{t1} & a_{t2} \end{bmatrix}^{\intercal}$ and $a_{f} = \begin{bmatrix} a_{f1} & a_{f2} \end{bmatrix}^{\intercal}$ describe the position of the moving contact, $\{ C \}$ or $\{ P \}$, on the tool's/finger's contacting surface. $v_{CP} = J \begin{bmatrix} \dot{u}_{t}^{\intercal} & \dot{u}_f^{\intercal} \end{bmatrix}^{\intercal}$ is $\{ P \}$'s Cartesian linear velocity relative to $\{ C \}$ for all contacts, and $J$ is the corresponding Jacobian \cite{math_intro_to_mani, modern_robotics}. $E_{xy}$ and $E_{z}$ are constant matrices extracting the $xy$- and $z$-elements of $v_{CP}$, respectively. $motion\_range \left( \cdots \right) \geq 0$ safeguards the bounds of $ua = \begin{bmatrix} ua_{t}^{\intercal} & ua_{f}^{\intercal} \end{bmatrix}^{\intercal}$ by predicting the next position in a forward Euler step. $collision\_linearized \left( \cdots \right) \geq 0$ is identical to the $collision \left( \cdots \right) \geq 0$ in hand sampling, but linearized. 

Physically, this QP asks the hand to hold the tool as steadily as possible while maintaining contact with the tool. After the QP computes $\dot{u}_f$, $\dot{a}_t$ and $\dot{a}_f$ are computed via 1st-order roll-slide contact kinematics \cite{montana_contact_kinematics, math_intro_to_mani}, and assembled into $\dot{ua}$. Then, an explicit 4th-order Runge-Kutta integrator integrates $\dot{ua}$ to obtain $ua$. The process continues until any of the following termination conditions is triggered: 1) The QP fails to converge. 2) A static equilibrium is impossible. 



We prescribe $\dot{u}_{t}$ as a constant and need not solve for it because the tool-hand parallel mechanism specifies the tool's path. In this way, the contacts act as joints whose kinematics is governed by the QP. The path planning is analogous to solving the tool-hand parallel mechanism's inverse kinematics, similar to backdriving an engine from its output shaft. When the shaft turns, other parts of the engine will move accordingly based on the engine's assembly. 

The above methods generate a dataset of all candidates' tool-wielding paths. Then, one can apply any metrics to the dataset to evaluate the candidates.

\section{Results \& Discussion}
\begin{figure*}[t]
    \centering
    \includegraphics[width=0.99 \textwidth]{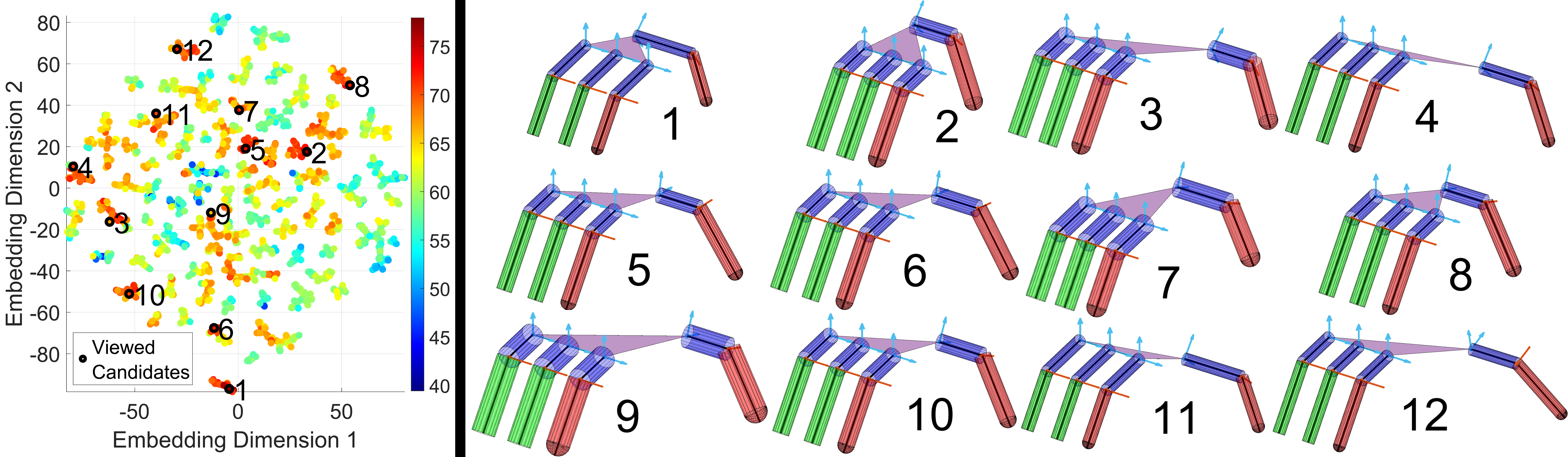}
    \caption{(Left) Mean score landscape of the 6D sampling case after dimensionality reduction with t-SNE. (Right) Hand design candidates in different high-performing clusters. The index next to each hand corresponds to a ``Viewed Candidates" index in the t-SNE plot. All hands are in the same scale. }
    \label{fig:results_6d}
\end{figure*}

\subsection{Sampling in a Two-Dimensional Design Space}

We first sampled the palm's length, $d_2$, and the palm's width, $d_3$, while keeping other design parameters constant at their initial values. In this way, we can visualize the results in 3D plots and gain intuition. We defined three metrics: 
\begin{enumerate}
    \item Tool motion range (TMR), defined as the tool's rotational magnitude along its path. 
    \item Mean contact sliding speed (MSS), defined as the mean of all contacts' sliding speeds along their paths. 
    \item Maximum finger torque (MFT), defined as the maximum joint torque required to withstand $10$ N of static force in the tool's cutting direction along a path.
\end{enumerate}
For metric 1), a larger value means better. For metrics 2) and 3), a smaller value means better. To facilitate comparisons, we normalized the metrics to dimensionless scores. For each metric, the worst performance was linearly mapped to a score of $0$ and the best to $100$. Three FPs and three metrics yield nine scores. Treating all nine scores as a cost vector, we computed the Pareto front \cite{matlab_pareto}. We also computed each candidate's mean score by averaging all nine scores. 

$785$ candidates, i.e., sampled hand designs that reached all FPs, were generated over $78,511$ RRT calls in this 2D design space before termination due to consistently low sampling efficiency at $1\%$. Shown in Fig. \ref{fig:results_eta} top left, as the sampling progressed, exponentially more samples were rejected. This was because the candidates gradually filled in the connected part of the feasible design space where the initial candidate was located, without spreading outside the space or overlapping with each other. Hence, more candidates mean less remaining room for new candidates. The standardized design space has a volume of $1$. If each candidate represents a circle with a radius of half minimum distance between any two candidates, the candidates would occupy about $68.50 \%$ of the volume. 

Fig. \ref{fig:results_eta} top shows how many of the candidates achieved a tool motion range above a variable threshold from each FP. Fig. \ref{fig:results_2d} shows the nine performance landscapes. We observed the following: 1) Based on tool motion range, all candidates succeeded in wielding the tool from all FPs. This verifies Hypothesis \ref{hpth:tool_hold} in this 2D case. 2) Candidates with similar performances cluster, and the performance landscapes have obvious contours and gradients. 3) The set of non-Pareto-optimal candidates also appear to cluster. Observations 2) and 3) reflect insights that could be difficult for methods that return a single ``optimal" design to uncover. 

\subsection{Sampling in the Entire Six-Dimensional Design Space}
We then sampled in the full 6D design space with identical parameters as were used in the 2D case. $10,000$ candidates were generated over $21,969$ RRT calls before termination due to reaching $10,000$ candidates and long computation time. The terminal sampling efficiency was $45.52 \%$, far above the termination threshold of $1 \%$ in the 2D case. Regarded as 6D balls, the $10,000$ candidates occupy $1.11 \times 10^{-4} \%$ of the 6D standardized design space's volume, about $10^{6}$ times lower than in the 2D case. This is because volume scales exponentially with dimensionality. Fig. \ref{fig:results_eta} top right shows the sampling efficiency trend. 

$99.42 \%$ of the $10,000$ candidates achieved greater-than-zero tool motion ranges for all FPs. The same ratios for carve, poke, and press individually are $99.42 \%, 100 \%$, and $100 \%$, respectively, as Fig. \ref{fig:results_eta} top right shows. These numbers support Hypothesis \ref{hpth:tool_hold}, while indicating that carve was the bottleneck. A closer look revealed that all failures were due to either unsatisfied static equilibrium once the fingers moved from the FP, or fingers reaching their joint limits at the FP. 

For visualization, we applied the t-distributed stochastic neighbor embedding (t-SNE) to transform the 6D design parameters $d$ into 2D embeddings. We chose t-SNE because it preserves distance relations among candidates. Fig. \ref{fig:results_6d} left shows the performance landscapes of mean score after we applied t-SNE. The candidates exhibit noticeable clustering, but are separated into isolated islands, some of which have obvious contours and gradients, whereas some do not. Meanwhile, though the candidates were enforced to sufficiently differ in the 6D design space, they overlap in the 2D embedding space, making them hard to distinguish. Nevertheless, candidates from different clusters exhibit clear differences, as shown in Fig. \ref{fig:results_6d} right. 

In total, the 6D and the 2D sampling cases produced $100,480$ hand designs. $10,785$ or $10.7\%$ of them reached all three FPs; more than $99\%$
of the $10,785$ hands that reached all three FPs successfully wielded the tool, supporting Hypothesis \ref{hpth:tool_hold}. Furthermore, the $10.7\%$ overall sampling efficiency and sampling efficiency trends in Fig \ref{fig:results_eta} top show that only a portion of the sampled hands reached all FPs. This indicates that the three FPs along with other conditions in the framework in Fig. \ref{fig:algorithm} are a strong but not overly restrictive selection mechanism thanks to RRT. If a less guided sampler is used, such as a completely random sampler, the sampling efficiencies could be lower. 


\subsection{Hardware Experiment}
We built the robotic hand hardware prototype based on the design parameters of the magenta point in Fig. \ref{fig:results_2d}. We chose this candidate for its convenient dimensions given the sizes of available hardware components---such as bearings---and because it is a Pareto-optimal design with reasonably good performances. To control the hand, we interpolated the joint position paths planned in the hand evaluation step with quintic trajectories, and applied a proportional-derivative (PD) controller to track the trajectories. During the experiment, the hand first moved to a trajectory's starting pose, then a human put the tool in the hand, and the hand executed the trajectory in an open loop. Additionally, the hand was equipped with enlarged soft skins to compensate for imprecision of control. 

We had the hand wield clay sculpting tools to manipulate a piece of Play-Doh \cite{play_doh}, which simulated clay. The hand managed to cut out a chunk of the clay and make marks on the clay with the tools. Besides, we discovered that, for carve, bracing with a 4th contact between the tool and the index finger's proximal link drastically improved the cutting performance, though this 4th contact was not considered during the design or the path planning. Fig. \ref{fig:results_eta} bottom shows snapshots of the experiment. 



\subsection{Limitations and Future Work}
We summarize the following limitations of this work: 1) Only three FPs were identified and implemented in this work. Hypothesis \ref{hpth:tool_hold}'s applicability toward more general tool-wielding and dexterous manipulation behaviors remains to be seen. 2) The candidates' performances depend on their FPs, but the optimization step that puts a hand into the FPs does not consider the evaluation metrics. This has led to candidates that reached the FPs but poorly wielded the tool, suggesting that factors beyond pose and statics should be considered in the optimization. 3) High-dimensional design spaces require enormous amounts of samples to explore, and their performance landscapes are hard to intuitively visualize. 

For future works, we plan to further study the clustering phenomena among the candidates, and identify families of high-performing designs. Meanwhile, we plan to investigate how to leverage the velocities and forces in tool-wielding behaviors to better design and control a hand. Lastly, we aspire to expanding the concept of FPs to more general dexterous manipulation tasks. On top of the tool-wielding results shown in this work, one type of tasks that could enable a hand to be more useful is tool-loading: To let the hand pick up a tool and configure itself into a specific FP with the tool, and to configure the hand out of the FP and put the tool away. We believe that the concept of FPs could be applicable to tool-loading as well, and we look forward to exploring this direction. 


\section{Conclusions}
This work's primary contribution is the idea that we can use \emph{foundational poses (FPs)} as a selection mechanism for the design of tool-wielding multi-finger robotic hands. If a hand can reach an FP, it will likely be able to wield the tool in a corresponding manner. This is because an FP distills a complex tool-wielding behavior into one pose that captures the working of an underlying tool-hand parallel mechanism. 

We tested this idea in a hand design experiment, for which we developed a sampling-based design optimization framework that uses three FPs we identified to constrain the feasible design space. The sampling uses RRT and starts from a hand that reaches all FPs to increase the chance of finding more hands that can reach the FPs. We sampled $100,480$ hand designs in total. $10,785$ of them reached all three FPs, and more than $99\%$ of the hands that reached all three FPs successfully wielded the tool, supporting the primary contribution. Meanwhile, the resultant performance landscapes unveiled clusters of high-performing designs, the Pareto front, contours, and gradients, which are insights into the non-convex, multi-objective hand design optimization problem that could be hard to uncover with methods that return one ``optimal" design. 

Furthermore, we built a hardware prototype with rigid endoskeleton and soft skin, and had it attempt tool-wielding tasks. With even open-loop joint position control, the prototype successfully executed the tool-wielding movements and manipulated a piece of clay with non-trivial interaction forces. The prototype achieved so even if this work only considered kinematics and statics. These results demonstrated our idea's real-world feasibility and potential. Lastly, we summarized this work's limitations and future works. We aspire to further investigate how the concept of FPs could advance the state-of-the-art of dexterous manipulation. 

\section{Acknowledgement}
This research was partially supported by NSF Convergence Accelerator award ITE-2344109, by the AI Research Institutes program supported by NSF and USDA-NIFA under AI Institute: for Resilient Agriculture, Award No. 2021-67021-35329, by the RAI Institute, and by the Technology Innovation Program (20018112, Development of autonomous manipulation and gripping technology using imitation learning based on visual-tactile sensing) funded by the Ministry of Trade, Industry \& Energy (MOTIE, Korea). 

\bibliographystyle{IEEEtran}
\bibliography{1st_Paper.bib}

\end{document}